\documentclass{article}

\usepackage{PRIMEarxiv}

\usepackage[utf8]{inputenc}
\usepackage[T1]{fontenc}
\usepackage[hidelinks]{hyperref}
\usepackage{url}
\usepackage{booktabs}
\usepackage{amsfonts}
\usepackage{nicefrac}
\usepackage{microtype}
\usepackage{lipsum}
\usepackage{fancyhdr}
\usepackage{graphicx}
\usepackage{tikz, pgfplots, pgfplotstable, schemabloc}
\usepackage{amsmath}

\usetikzlibrary{shapes,shapes.geometric,arrows,fit,calc,positioning,automata,patterns}
\graphicspath{{media/}}
\definecolor{rosso}{RGB}{220,57,18}
\definecolor{giallo}{RGB}{255,153,0}
\definecolor{blu}{RGB}{102,140,217}
\definecolor{verde}{RGB}{16,150,24}
\definecolor{viola}{RGB}{153,0,153}
\definecolor{ggrigio}{RGB}{200,200,200}
\definecolor{bblue}{HTML}{4F81BD}
\definecolor{rred}{HTML}{C0504D}
\definecolor{ggreen}{HTML}{9BBB59}
\definecolor{ppurple}{HTML}{9F4C7C}
\definecolor{palette_blue}{HTML}{20639B}
\newcolumntype{L}[1]{>{\raggedright\arraybackslash}p{#1}}
\newcolumntype{C}[1]{>{\centering\arraybackslash}p{#1}}
\newcolumntype{R}[1]{>{\raggedleft\arraybackslash}p{#1}}
\title{An adaptable cognitive microcontroller node for fitness activity recognition
}

\author{
  Matteo Antonio Scrugli \\
  Department of Electrical and Electronic Engineering (DIEE) \\
  University of Cagliari \\
  Cagliari, Italy \\
  \texttt{matteo.scrugli@unica.it} \\
  \And
  Bojan Blažica \\
  Computer Systems Department \\
  Jožef Stefan Institute \\
  Ljubljana, Slovenia \\
  \texttt{bojan.blazica@gmail.com} \\
  \And
  Paolo Meloni \\
  Department of Electrical and Electronic Engineering (DIEE) \\
  University of Cagliari \\
  Cagliari, Italy \\
  \texttt{paolo.meloni@unica.it} \\
}

\begin{document}
\maketitle

\begin{abstract}
The new generation of wireless technologies, fitness trackers, and devices with embedded sensors can have a big impact on healthcare systems and quality of life. Among the most crucial aspects to consider in these devices are the accuracy of the data produced and power consumption.
Many of the events that can be monitored, while apparently simple, may not be easily detectable and recognizable by devices equipped with embedded sensors, especially on devices with low computing capabilities. It is well known that deep learning reduces the study of features that contribute to the recognition of the different target classes.
In this work, we present a portable and battery-powered microcontroller-based device applicable to a wobble board. Wobble boards are low-cost equipment that can be used for sensorimotor training to avoid ankle injuries or as part of the rehabilitation process after an injury. The exercise recognition process was implemented through the use of cognitive techniques based on deep learning.
To reduce power consumption, we add an adaptivity layer that dynamically manages the device's hardware and software configuration to adapt it to the required operating mode at runtime. Our experimental results show that adjusting the node configuration to the workload at runtime can save up to 60\% of the power consumed. On a custom dataset, our optimized and quantized neural network achieves an accuracy value greater than 97\% for detecting some specific physical exercises on a wobble board.
\end{abstract}


\section{Introduction}
\label{sec:introduction}
The guidelines of the World Health Organization (WHO) in 2010 document, excluding special cases, an average adult should engage in physical activity of moderate intensity for at least 150 minutes per week and 75 minutes per week at high intensity\cite{who}. Tracking and encouraging good levels of physical activity can improve people's health.\cite{intro1}
Fitness tracker devices have had a rapid development in recent years, due to their ease of use, accuracy, and portability. 
Events in a trackable signal, although seemingly simple, can be difficult to identify and recognise within the data stream.
Deep learning is well known for reducing the study of features that contribute to the recognition of different target classes and greatly increase the classification accuracy, but in order to be used in low power devices a careful software optimization is necessary.

In this work, we present a portable and battery-powered microcontroller-based device applicable to a wobble board. Wobble boards are inexpensive and easy-to-use tools to avoid ankle injuries or as part of the recovery process after an injury. The exercise recognition process was implemented through the use of cognitive techniques based on deep learning.
To manage the hardware/software configuration we have implemented a component called ADAM (ADAptive runtime Manager), able to optimize device power consumption and performance.
ADAM creates and manages a network of processes that communicate with one another via FIFOs. The morphology of the process network varies depending on the operating mode (OM) in execution.
ADAM can be triggered by external environment re-configuration messages or by specific workload-related variables in the sampled streams. When triggered, ADAM alters the morphology of the process network by turning on or off processes and rearranging the inter-process FIFOs.
Furthermore, depending on the new configuration, it modifies the platform's hardware configuration, adjusting power-related settings such as clock frequency, supply voltage, and peripheral gating.
\section{Related work}
\label{sec:relatedwork}
Local processing is frequently used only for implementing simple checks on raw data and/or marshaling tasks for wrapping sensed data inside standard communication protocols, and the edge-computing paradigm is only marginally exploited\cite{lp1,lp3,lp4}.
More complex and accurate algorithms, such as those based on artificial intelligence or deep learning, must be targeted in order to properly use cognitive computing at the edge. Their effectiveness on high-performance computing platforms has been widely demonstrated.
Nonetheless, how to map state-of-the-art cognitive computing on resource-constrained platforms remains an open question. To identify specific events in sensed data, an increasing number of approaches based on machine learning and artificial intelligence are being developed.
In \cite{nodata} and \cite{bologna} ANN (artificial neural networks) are used by the authors to detect specific conditions in the proposed data. In \cite{bologna}, an ANN is used to determine the patient's emotional state (happiness or sadness).
In \cite{prev}, energy/power efficiency is improved, using near-sensor processing to save data transfers, and dynamically adapting application setup and system frequency to the OM requested by an external user and to data-dependent workload.

In our use case, a CNN (Convolutional Neural Network) is used to identify and recognize simple physical exercises performed on a wobble board.
In \cite{game,bojan}, the authors propose the use of a wobble board in creative way, to entice people to its use, as it is very important in ankle rehabilitation. 
In a review of the concept of patient motivation \cite{mc}, 
the authors describe how motivation has been considered in relation to rehabilitation associated with strokes, fractures, rheumatic disease, aging, and cardiac and neurological issues. The limited motivation on the part of some individuals may, at least in part, be ascribed to the tedious nature of the ankle exercises and the inability to monitor one’s improvement throughout the course of the training process \cite{exercises}.

A similar approach is considered in our work, the implemented system detects and identifies some simple sensorimotor exercises performed on the wobble board, giving a percentage of correct execution at the end of the exercise. As far as we know, our system is the only one that applies state-of-the-art deep learning-based techniques to recognize some specific movements on a wobble board, managing hardware and software dynamically in order to minimize power consumption.
\section{Wobble board \& node architecture}
\label{sec:node}
\begin{figure}
    \centering
    \includegraphics[width=0.65\textwidth]{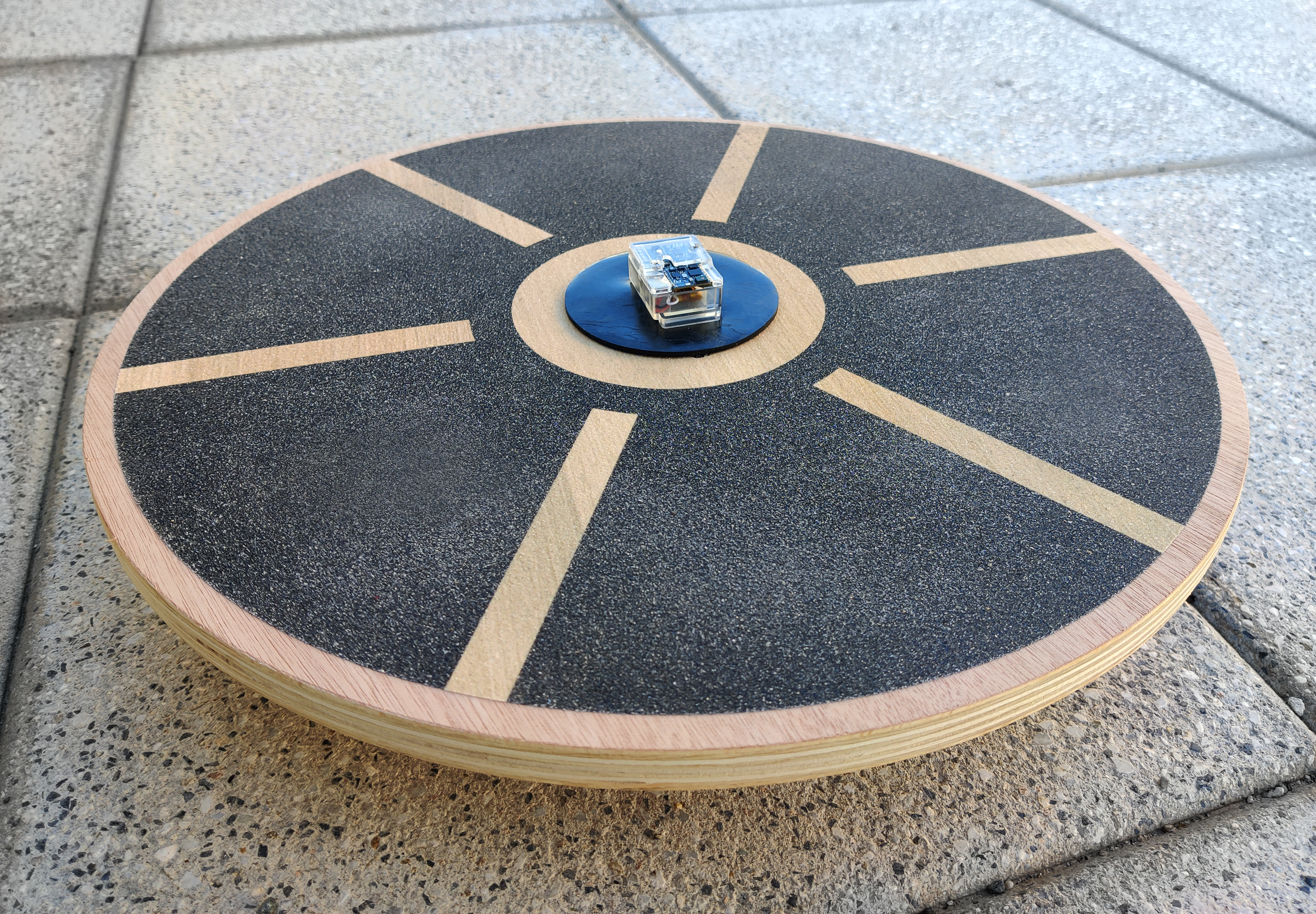}
    \caption{Wobble board used to validate our approach.}
    \label{fig:board}
\end{figure}

We used a wobble board capable of 360° rotation (Figure \ref{fig:board}), the sensory node is fixed in the upper-middle part. Since the device is battery powered and communication is via a Bluetooth Low Energy module, no cable is used to interface with the sensor node.
We chose STMicroelectronics SensorTile microcontroller-based platform, which is equipped with an ARM Cortex-M4 32-bit low-power microcontroller. It takes advantage of the LSM303AGR accelerometer sensor integrated into the Sensortile, only the two axes X and Y parallel to the floor are taken into account.
It was chosen to run FreeRTOS on the node, to have more control over the running tasks due to its ability to create a thread-level abstraction.

\subsection{Application model}
\label{sec:app_model}
We chose a process network-based application structure. Tasks are modeled as separate processes that communicate with one another via FIFO structures. Using a software pipeline, processes can potentially be executed in parallel, improving performance. When the topology of the network processes changes, a change of OM occurs.

We identified four topologies of processes that can be combined in different ways:

\noindent
$\bullet$\quad\textit{\textbf{Get data task}:} take data from the sensing hardware.

\noindent
$\bullet$\quad\textit{\textbf{Process task:}} it's possible to have multiple tasks of this type, representing multiple stages of in-place data analysis algorithm.

\noindent
$\bullet$\quad\textit{\textbf{Threshold task:}} filters data depending on the results of the analysis.

\noindent
$\bullet$\quad\textit{\textbf{Send task:}} is the task in charge of outwards communication to the gateway.

\subsection{Adaptivity support: the ADAptive runtime Manager}
\label{sec:adam}
A task within the process network was dedicated to the management of the platform's dynamic hardware and software reconfiguration. We have implemented such reconfiguration in a software agent called ADAptive runtime Manager (ADAM). ADAM can be activated on a regular basis by using an internal timer, it monitors the system's status, such as changes in workload. ADAM can react to such input by changing the platform settings, performing various operations such as enabling or disabling individual tasks of the sensor task chain or the entire chain; deciding whether to put the microcontroller in sleep mode or not; setting the operating frequency; and rerouting the data-flow managed by the FIFOs based on the active tasks.
\section{Designing the application}
\label{sec:usecase}
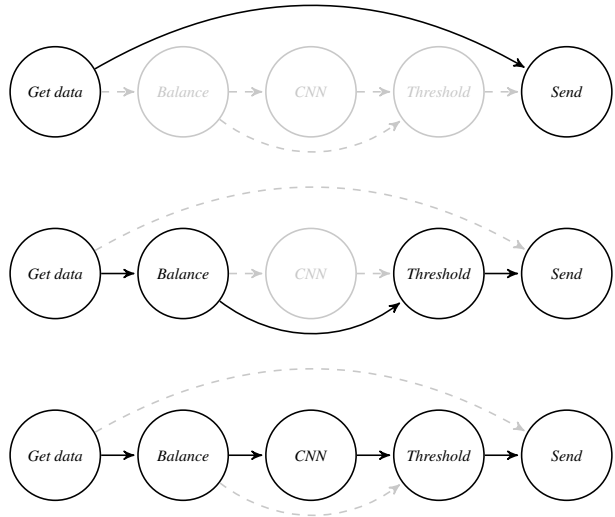
\begin{figure}
	\tiny
	\centering
	\begin{tikzpicture}[->,>=stealth',initial text={},shorten >=1pt,auto,node distance=1.7cm,semithick]
	\node
	    [state, minimum size=1.2cm] (x0)					                          {\textit{Get data}};
	\node
	    [state, minimum size=1.2cm, draw=ggrigio, text=ggrigio] (x1)
	    [right of=x0]
	    {\textit{Balance}};
	\node
	    [state, minimum size=1.2cm, draw=ggrigio, text=ggrigio] (x2)
	    [right of=x1]
	    {\textit{CNN}};
	\node
	    [state, minimum size=1.2cm, draw=ggrigio, text=ggrigio] (x3)
	    [right of=x2]
	    {\textit{Threshold}};
	\node
	    [state, minimum size=1.2cm] (x4)
	    [right of=x3]
	    {\textit{Send}};
	
	\path
	(x0)	edge [ggrigio, dashed]	node {\textit{$ $}}	(x1)
	    	edge [bend left=30]                     node {\textit{$ $}} (x4)
	(x1)	edge [ggrigio, dashed]		            node {\textit{$ $}}	(x2)
	        edge [bend right=37, ggrigio, dashed]   node {\textit{$ $}} (x3)
	(x2)	edge [ggrigio, dashed]		            node {\textit{$ $}}	(x3)
	(x3)	edge [ggrigio, dashed]                  node {\textit{$ $}}	(x4);
	\end{tikzpicture}
	\qquad
	\begin{tikzpicture}[->,>=stealth',initial text={},shorten >=1pt,auto,node distance=1.7cm,semithick]
	\node
	    [state, minimum size=1.2cm] (x0)					                          {\textit{Get data}};
	\node
	    [state, minimum size=1.2cm] (x1)
	    [right of=x0]
	    {\textit{Balance}};
	\node
	    [state, minimum size=1.2cm, draw=ggrigio, text=ggrigio] (x2)
	    [right of=x1]
	    {\textit{CNN}};
	\node
	    [state, minimum size=1.2cm] (x3)
	    [right of=x2]
	    {\textit{Threshold}};
	\node
	    [state, minimum size=1.2cm] (x4)
	    [right of=x3]
	    {\textit{Send}};
	
	\path
	(x0)	edge                                    node {\textit{$ $}}	(x1)
	    	edge [bend left=30, ggrigio, dashed]    node {\textit{$ $}} (x4)
	(x1)	edge  [ggrigio, dashed]   		        node {\textit{$ $}}	(x2)
	        edge [bend right=37]                    node {\textit{$ $}} (x3)
	(x2)	edge  [ggrigio, dashed]   		        node {\textit{$ $}}	(x3)
	(x3)	edge                                    node {\textit{$ $}}	(x4);
	\end{tikzpicture}
	
	\begin{tikzpicture}[->,>=stealth',initial text={},shorten >=1pt,auto,node distance=1.7cm,semithick]
	\node
	    [state, minimum size=1.2cm] (x0)					{\textit{Get data}};
	\node
	    [state, minimum size=1.2cm] (x1)
	    [right of=x0]
	    {\textit{Balance}};
	\node
	    [state, minimum size=1.2cm] (x2)
	    [right of=x1]
	    {\textit{CNN}};
	\node
	    [state, minimum size=1.2cm] (x3)
	    [right of=x2]
	    {\textit{Threshold}};
	\node
	    [state, minimum size=1.2cm] (x4)
	    [right of=x3]
	    {\textit{Send}};

	\path
	(x0)	edge                                    node {\textit{$ $}}	(x1)
	    	edge [bend left=30, ggrigio, dashed]    node {\textit{$ $}} (x4)
	(x1)	edge                    		        node {\textit{$ $}}	(x2)
	        edge [bend right=37, ggrigio, dashed]   node {\textit{$ $}} (x3)
	(x2)	edge                                    node {\textit{$ $}}	(x3)
	(x3)	edge                                    node {\textit{$ $}}	(x4);
	\end{tikzpicture}
	
	\caption{Application model. Top \textit{raw OM}, middle \textit{balance OM}, bottom \textit{CNN OM}.}
	\label{fig:opmodeconf}
\end{figure}

In this work, a system is implemented that is able to recognize typical movements in exercises that involve the use of a conventional wobble board (described in Section \ref{sec:node}) or, more simply, the wireless transmission of raw data acquired from the sensor. The application model chosen for this use case provides two possible levels of processing able to evaluate the nature of the movement. The OMs chosen for the selected use case are shown in Figure \ref{fig:opmodeconf} and described below.

\subsection{Operating mode: raw data}
This is the simplest OM, using only two tasks. It is possible to acquire data from the sensor and send it via Bluetooth, with a sampling frequency of 100 Hz. In order to reduce the power consumption related to the transmission, it was decided to encapsulate four samples taken from the sensor in a single low energy Bluetooth packet. The Bluetooth packet has a size of 20 bytes, 4 bytes the timestamp, and four pairs of data taken from the sensor at different instants of time, the data pair is formed by the values relative to the accelerometer's X and Y axis, each with a size of 2 bytes.

\subsection{Operating mode: basic balance}
This OM enables the first level of processing. The sampling rate is lowered to 100/7 Hz, which is more than sufficient to perform the analysis in this OM. A simple algorithm calculates how much, in percentage, the wobble board is in a balanced position.
The extreme cases, the analysis returns a value of 100\% if the board remains horizontal within a certain tolerance and 0\% when the board remains in constant contact with the ground.
The result of the analysis is transmitted every second, this leads to a significant energy saving due to the decrease of information that has to be sent via Bluetooth, which is no longer used to transmit raw data. 

\subsection{Operating mode: CNN}
\begin{figure}
    \centering
    \includegraphics[width=0.95\textwidth]{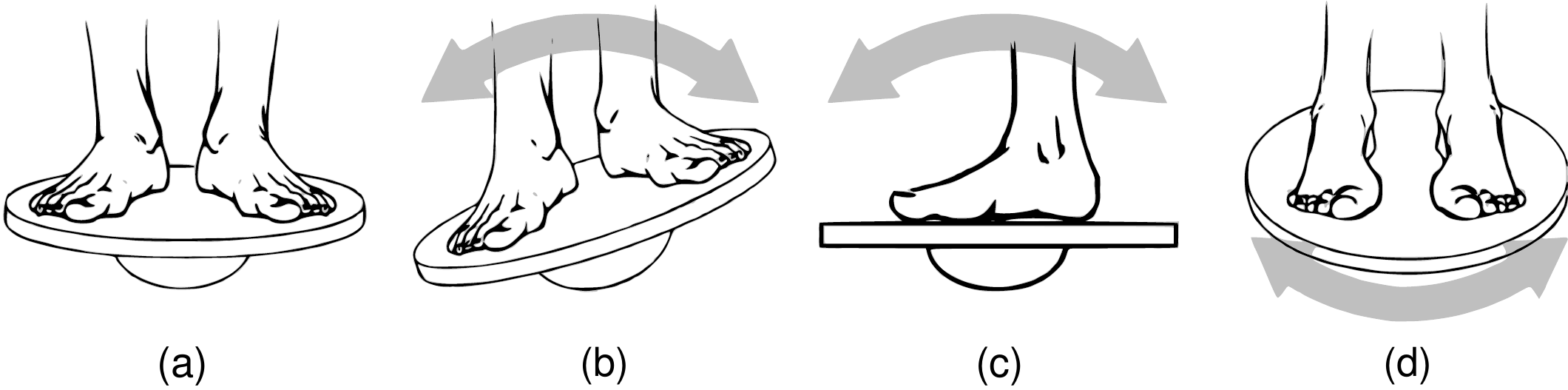}
    \caption{The four common wobble board exercises recommended by physiotherapist Anders Heckmann \cite{exercises}. (a) Balance while keeping as steady as possible. (b) Move the board back and forth. (c) Move the board from sideto side. (d) Clockwise and counterclockwise circular movement. The Figure was extracted from \cite{game}}
    \label{fig:exe}
\end{figure}

Some exercises were selected which were not too complicated to be recognized by deep learning techniques. Also in this case, a frequency of 100/7 Hz is ideal to obtain good results with the neural network and have a not excessive workload. The raw data related to the two X and Y axes of the accelerometer are used as two different input features as input to the neural network.
The \textit{balance} task remains active so that if a total stop of the table is detected, no CNN is executed. Again, the result of the analysis is transmitted every second.
Some typical exercises recommended by Anders Heckmann \cite{exercises} are those shown in Figure \ref{fig:exe}.

\noindent The correct execution of the exercise involves:
\begin{itemize}
    \item \textbf{Basic stance balance (Figure \ref{fig:exe}.a):} Stand on the board with the edges of your feet on the outer edges of the board. Maintain a neutral spine and keep your torso upright. Balance on the board by shifting your weight to prevent any of the board’s edges from touching the floor. The goal is to maintain the balance for 60 seconds.
    \item \textbf{Forward/backward tilt (Figure \ref{fig:exe}.b):} Stand on the wobble board with your feet on the outer edges. Stand upright in a neutral spine position. Tilt the board to the front to touch the floor. Tilt it back onto the heels to touch the floor behind you. Continue tilting forward and back in a slow, steady, controlled motion for 60 seconds.
    \item \textbf{Side tilt (Figure \ref{fig:exe}.c):} Stand on the wobble board with your feet on the outer edges. Stand upright in a neutral spine position. Tilt the board from left to right by transferring your weight from your left leg to your right leg. Moving in a slow and controlled manner, keeping an upright torso and tight core. The duration of the exercise is 60 seconds.
    \item \textbf{Two leg tilts (Figure \ref{fig:exe}.d):} Stand on the wobble board with your feet on the outer edges. Stand upright in a neutral spine position. In a combination of the two previous exercises, you will roll the board in a 360-degree motion. Begin by tilting the board to the left. When the board touches the ground on the left, transfer your weight to the front to touch the floor. Now transfer your weight to touch the floor to the right side. Complete the revolution by tilting the board to the floor behind you. Keep your body centralized throughout. You may need to balance with your arms as you get used to the movement. Reverse the motion to move in the other direction. Continue for 60 seconds.
    \item \textbf{Other:} there is a fifth class that represents everything that is not foreseen by the previous exercises, for example the fall from the table or the absolute absence of movement.
\end{itemize}

\subsection{Neural network design}
We used a training procedure that included a static quantization\footnote{\label{note:quant}\url{https://pytorch.org/tutorials/advanced/static_quantization_tutorial.html}} step, the source code is available in our public repository\footnote{\url{https://github.com/matteoscrugli/deepwobbleboard}}. This process converts floating-point weights and activations to integers, allowing the CNN to be implemented using the CMSIS-NN optimized function library, which expects inputs with 8-bit precision.
We have chosen to force the value of the bias to zero, while for the conversion of the weights we have inserted \textit{MinMax} observers\footnote{\url{https://pytorch.org/docs/stable/_modules/torch/quantization/observer.html}}, who have the task of studying the outputs of each layer. Evaluating the distribution of the output values of each layer allows the observer to establish a value of \textit{scale} and \textit{zero-point} in order not to saturate these values using a quantized network.
The CMSIS-NN library's functions for implementing convolution and fully connected layers include output shifting operations for applying the Scale factor to the outputs, with scaling values ranging from -128 to 127. In PyTorch, however, the quantization procedure requires a Scale value that is not always a power of two. As a result, we modified the CMSIS functions slightly to support arbitrary \textit{scale} values. This change resulted in a minor increase in inference execution time. After testing inference with and without this modification, we calculated an increase in execution time of 2.87\%.

\begin{figure}
    \centering
    \begin{minipage}{0.9\textwidth}
        \centering
        \includegraphics[width=1\textwidth]{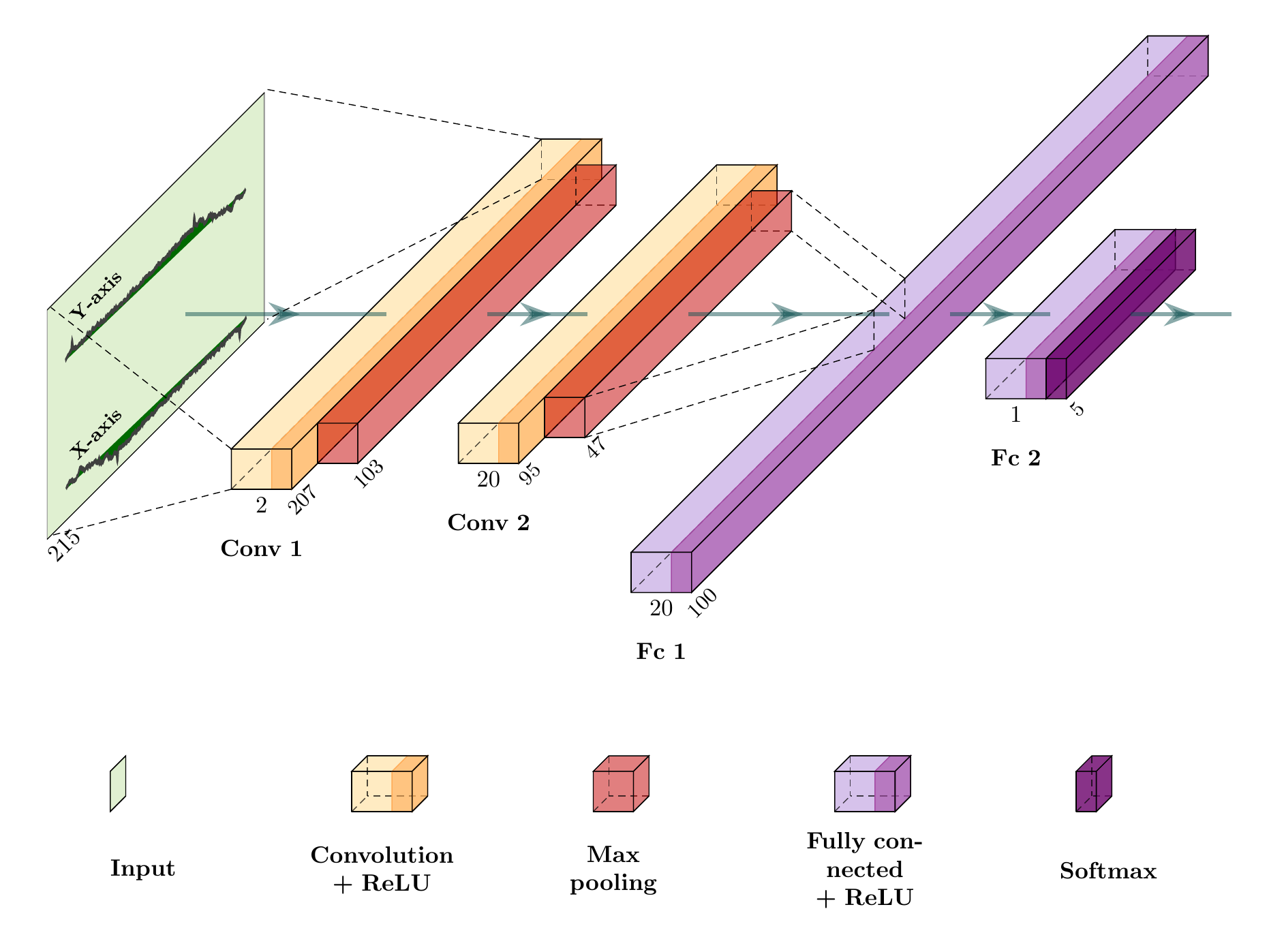}
    \end{minipage}
    \\[2.5mm]
    \hfill
    \begin{minipage}{1\textwidth}
        \centering
	\footnotesize
	\begin{tabular}{L{48mm} L{48mm}}
		\hline
		\\[-2.5mm]
		\textbf{(B)} Basic stance balance & \textbf{(FB)} Forward/backward tilt\\
		\\[-2.5mm]
		\textbf{(S)} Side tilt & \textbf{(R)} Two leg tilts \\
		\\[-2.5mm]
		\textbf{(G)} Other\\
		\\[-2.5mm]
		\hline
	\end{tabular}
    \end{minipage}
  \caption{CNN structure and classes description.}
  \label{fig:cnn}
\end{figure}

We used a design space exploration process to compare tens of neural network topologies in terms of accuracy achieved after training and computing workload associated with executing the inference task on SensorTile. Figure \ref{fig:cnn} shows the selected convolutional network and the five selected output classes.
    
\noindent All exercises are one minute in length, each movement performed during the four different exercises has a different duration. Generally, the longest exercise is the two leg tilts. CNN does not analyze the exercise for its entire duration (60 seconds) at once, the signal is divided into windows of 15 seconds duration, a good compromise between temporal precision and distinction between the movements to be evaluated. The maximum number of epochs was set to 30 and the Early Stopping (ES) algorithm was chosen to avoid overfitting effects. This algorithm terminates the training phase if it detects an increase in the loss value\cite{goodfellow}; the loss is evaluated every epoch, and a Patience value of 5 is selected, implying that the training terminates only if a loss increment is detected for 5 consecutive epochs. Table \ref{tab:hp} summarizes the parameters chosen for training, while Table \ref{tab:cnn}, shows the dimensions of the various layers chosen.

\begin{table}
	\centering
	\footnotesize
	\begin{tabular}{C{25mm} C{25mm} C{8mm} C{25mm} C{25mm}}
		\hline
		\\[-2.5mm]
		\textit{\textbf{Hyperparameter}} & \textit{\textbf{Value}} & &
			\textit{\textbf{Hyperparameter}} & \textit{\textbf{Value}}\\
		\\[-2.5mm]
		\hline
		\hline

		\\[-2.5mm]
		Epochs & 30 & & Optimizer & Adadelta \\
		Batch size & 32 & & Learning rate & 1.0 \\
		Loss criterion & Cross Entropy & & Rho & 0.9 \\
		ES patience & 5 & & ES evaluation & Every epoch \\
		\\[-2.5mm]
		
		\hline
		\\[-1.5mm]
	\end{tabular}
	\caption{Hyperparameters used during the training phase.}
	\label{tab:hp}
\end{table}
\begin{table}
	\centering
	\footnotesize
	\begin{tabular}{C{24mm} C{13mm} C{13mm} C{13mm} C{13mm} C{13mm}}
		\\[0.5mm]
		\hline
		\\[-2.5mm]
		\textit{\textbf{Layer}} &
		\textit{\textbf{Input dimension}} &
		\textit{\textbf{Output dimension}} &
		\textit{\textbf{Input features}} &
		\textit{\textbf{Output features}} &
		\textit{\textbf{Kernel size}}
		\\
		\\[-2.5mm]
		\hline
		\hline
		
		\\[-2.5mm]
		Convolutional &     $ 215 $ &   $ 207 $ &   $ 2 $ &     $ 20 $ &    $ 9 $ \\
		Max pooling &       $ 207 $ &   $ 103 $ &   $ 20 $ &    $ 20 $ &    $ 2 $ \\
		Convolutional &     $ 103 $ &   $ 95 $ &    $ 20 $ &    $ 20 $ &    $ 9 $ \\
		Max pooling &       $ 95 $ &    $ 47 $ &    $ 20 $ &    $ 20 $ &    $ 2 $ \\
		Fully connected &   $ 940 $ &   $ 100 $ &   $ - $ &     $ - $ &     $ - $ \\
		Fully connected &   $ 100 $ &   $ 5 $ &     $ - $ &     $ - $ &     $ - $ \\
		\\[-2.5mm]
	
		\hline
        \\[-1.5mm]
	\end{tabular}
	\caption{Model parameters.}
	\label{tab:cnn}
\end{table}
\subsection{Data augmentation \& generalization}
Unfortunately, no datasets already used in the literature containing the selected exercises were found. Therefore, a dataset was created with 12 one-minute recordings for each type of exercise (including class ``other'').
A random split of the dataset was chosen in order to use 0.8\% of the data for the training set and 0.2\% for the validation set.
Operations such as translation, rotation and time dilation of the signal in each direction can often greatly improve generalization\cite{goodfellow}.

In more detail, augmentation techniques are (also summarized in Table \ref{tab:aug}):
\begin{itemize}
    \item \textbf{Translation:} During training, the window to the entire signal is shifted by 0.25 s per frame. For example, a 60-second recording with a translation of 0.25 generates a number of frames of $(60 - 15) / 0.25 + 1 = 181$.
    \item \textbf{Rotation:} A rotation transformation was applied to the X and Y axes of the sensor data, in our case we chose two rotations of $\angle{-4}$ and $\angle{4}$ degrees. For each record, two more are then generated.
    \item \textbf{Dilation:} The sampling frequency of the signals in the dataset is 100 Hz, but the neural network is trained with 100/7 Hz signals. The size of the input signal is therefore equal to $\lfloor(15 \times 100 + 7 - 1) / 7\rfloor = 215$. Time dilation can be obtained by increasing or decreasing the downsampling while keeping the input size to the neural network constant, in this case, two additional downsampling values of 6 and 8 were chosen. For each recording, two more are generated with different time dilations.
\end{itemize}

\begin{table}
	\centering
	\footnotesize
	\begin{tabular}{C{65mm} C{35mm}}
		\hline
		\\[-2.5mm]
		\textit{\textbf{Parameter}} & \textit{\textbf{Value}}\\
		\\[-2.5mm]
		\hline
		\hline

		\\[-2.5mm]
        Traslation, temporal distance between frames & 0.25 s \\
        Rotation, X and Y axis rotation & $\angle{-4}$, $\angle{0}$, $\angle{4}$\\
		Time dilation, downsampling & 6, 7, 8\\
		\\[-2.5mm]
		
		\hline
		\\[-1.5mm]
	\end{tabular}
	\caption{Augmentation parameters.}
	\label{tab:aug}
\end{table}
\section{Experimental results}
\label{sec:results}
In this section, we will show the results obtained after the neural network training and we will make a detailed analysis of the power consumption for each OM.

\subsection{Neural network accuracy}
\begin{figure}
    \centering
    \includegraphics[width=0.82\textwidth]{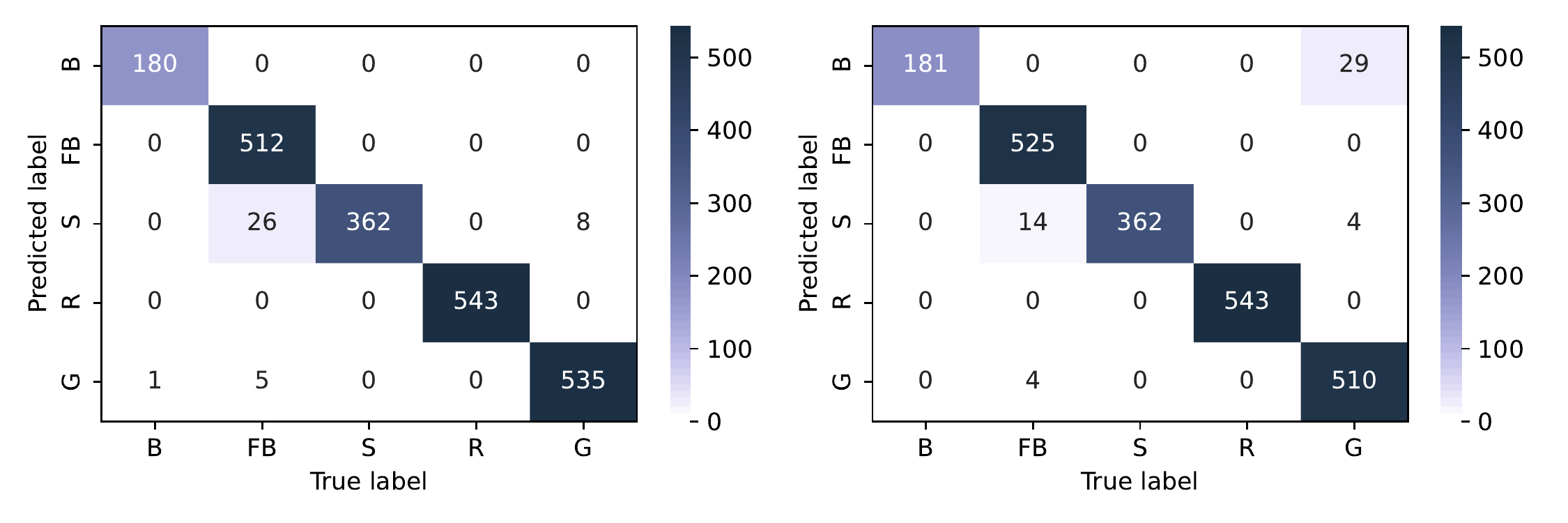}
    \caption{Validation set confusion matrix, to the left the model with floating point weights and to the right fixed point weights.}
  \label{fig:cm}
\end{figure}

After the training phase, an accuracy of 97.652\% was measured on the validation set.
Figure \ref{fig:cm} shows the results of the training, showing how the windows extracted from the validation set are classified. It is possible to notice that the major difficulty for the network is to recognize in a correct way when the wobble board is used with movements that do not match the four proposed ones.

\subsection{Power consumption}
We measured the power consumption for each OM, for this purpose 
the digital oscilloscope ANALOG Discovery 2 was used to measure the voltage on the shunt resistor placed in series to the power cable of the SensorTile node. Figure \ref{fig:ene} shows the result of the measurement.

\subsubsection{Operating mode \textit{raw}}
This is the OM with the highest amount of data to be sent via Bluetooth, the minimum system frequency to handle data traffic with the Bluetooth module present in the SensorTile module is 8 MHz. In order to optimize data sending via Bluetooth, four sensor acquisitions are merged for each packet, reducing the data sending frequency from 100 Hz to 25 Hz.

\subsubsection{Operating mode \textit{balance}}
\begin{figure}
    \centering
    \includegraphics[width=0.95\textwidth]{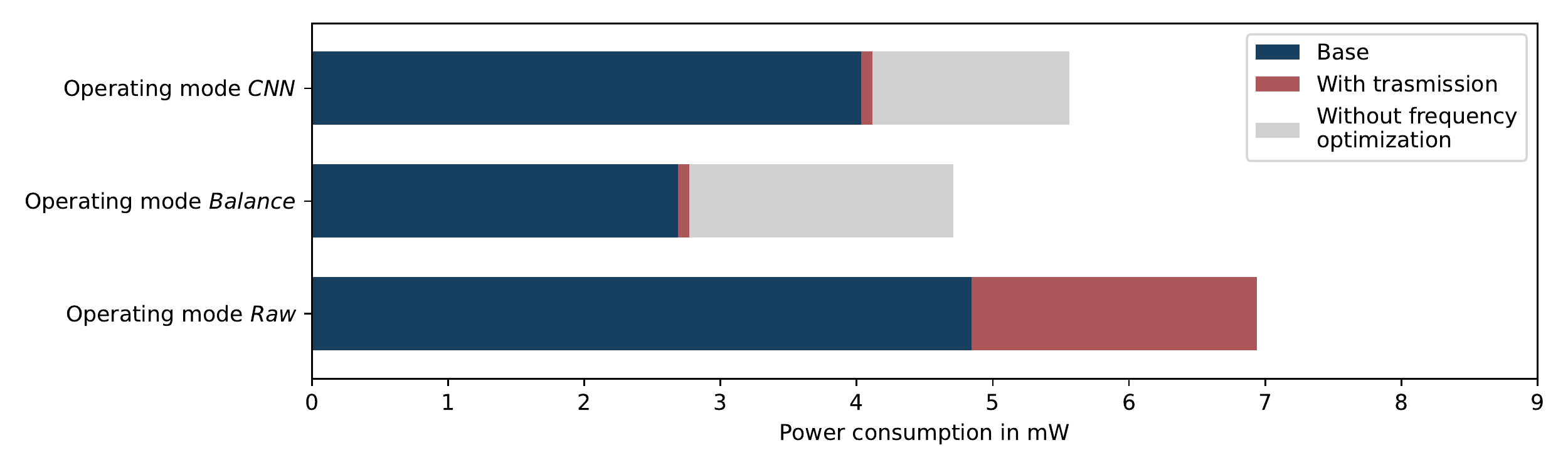}
    \caption{Power consumption for each OMs.}
    \label{fig:ene}
\end{figure}

In contrast to the previous one, this is the OM where there is less data transmission, in fact, the evaluation of the exercise is done every one second, invoking Bluetooth transmission at the same frequency. It has been tested that a system frequency of 2 MHz is sufficient to meet the real-time constraints, Figure \ref{fig:ene} shows the savings due to dynamic optimization of the system frequency.

\subsubsection{Operating mode \textit{CNN}}
It was chosen to send the information about the classification result of the exercise every time the neural network inference is performed. In order to correctly execute the neural network and at the same time respect the real-time constraints, a system frequency of 4 MHz has been set. The length of the input frame is obviously the same as that used during training, while the distance between frames in this evaluation phase, as for operating mode \textit{Balance}, is one second. For this reason, the power consumption is data-dependent and the worst case will thus be taken into account for the calculation of power consumption. The maximum number of times a single data item is sent via the Bluetooth module is equal to $(60 - 15) / 1 + 1 = 46$.

\subsection{Power consumption model}
We conducted a comprehensive set of experiments measuring energy consumption in various setup conditions. The results were used to create a model that highlighted the contribution of each task to the node's energy consumption.
The energy values for each task in the process network are shown in Table \ref{tab:taskCon}, Table \ref{tab:platCon} instead shows the power consumption of the platform as a function of the chosen system frequency

\begin{figure}
    \centering
    \includegraphics[width=1\textwidth]{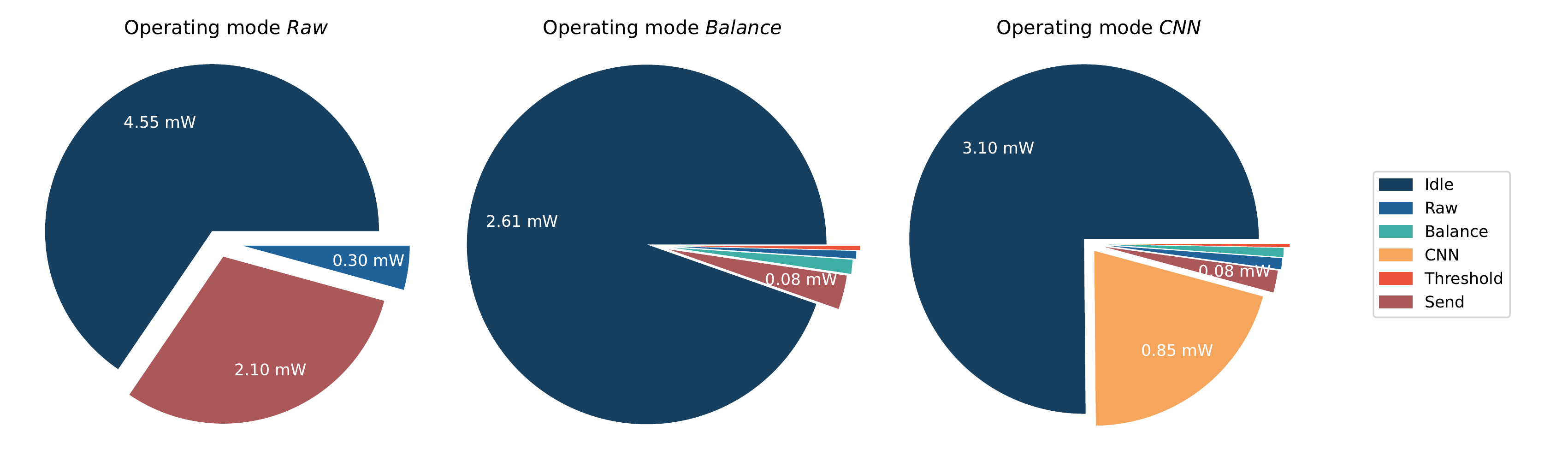}
    \caption{Estimation of energy consumption for each task of each OM.}
    \label{fig:pie}
\end{figure}

\begin{table}
	\centering
	\footnotesize
	\begin{tabular}{C{35mm} C{25mm} C{25mm} C{25mm}}
		\hline
		\\[-1.6mm]
		\textit{\textbf{Task type}} & \textit{\textbf{Number of cycles}} & \textit{\textbf{Execution time (\boldmath$8\,$MHz)}} & \textit{\textbf{Energy contribution}} \\
		\\[-3mm]
		\hline
		\hline
		\\[-2.5mm]
		\textit{\textbf{Get data}}      & $ 841 $       & $ 105 \,\mu s $   & $E_g = 2.96\,\mu J$   \\
		\\[-2.5mm]
		\textit{\textbf{Get data + balance}}          & $ 1\,550 + 841 $       & $ 300 \,\mu s $    & $ E_{gb} = 3.76\,\mu J$   \\
		\\[-2.5mm]
		\textit{\textbf{CNN}}    & $ 2\,219\,582 $    & $  277\,ms $      & $ E_c = 852.38\,\mu J$ \\
		\\[-2.5mm]
		
		\textit{\textbf{Threshold}}     & $ 910 $       & $ 114 \,\mu s $   & $ E_t = 2.73\,\mu J$    \\
		\\[-2.5mm]
		\textit{\textbf{Send data}}     & $ \sim25\,000 $ & $ \sim3 \,ms $    & $ E_s = 83.96\,\mu J$  \\
        \\[-2.5mm]
		\hline
        \\[-1.5mm]
	\end{tabular}
	\caption{Summary of consumption and execution time for each task.}
	\label{tab:taskCon}
\end{table}
\begin{table}
	\centering
	\footnotesize
	\begin{tabular}{C{37mm} C{13mm} C{13mm} C{13mm}}
		\hline
		\\[-1.6mm]
		\textit{\textbf{Device}} & \multicolumn{3}{c}{\textit{\textbf{\shortstack{Power consumption}}}}\\
		\cmidrule(lr){2-4} &
		\boldmath$2\,M\!H\!z$ & \boldmath$4\,M\!H\!z$ & \boldmath$8\,M\!H\!z$ \\
		\\[-3mm]
		\hline
		\hline
		\\[-2.5mm]
		\textit{\textbf{Platform in idle state}}  & $ 2.609\,mW $ & $ 3.101\,mW $ & $ 4.546\,mW $ \\
		\\[-2.5mm]
		\hline
        \\[-1.5mm]
	\end{tabular}
	\caption{Summary of consumption of peripherals.}
	\label{tab:platCon}
\end{table}
It is then possible to obtain the equations that estimate the consumption for each OM:
\begin{align}
    \label{eq:power1}
    P_{\textit{raw\,data\,OM}} &= (E_g+ \alpha E_{s}) \cdot f_s + P_{idle}\,,\\
    \label{eq:power2}
    P_{\textit{basic\,balance\,OM}} &= E_{gb} \cdot f_s + (E_t + E_s) \cdot f_{b} + P_{idle}\,,\\
    \label{eq:power3}
    P_{\textit{cnn\,processing\,OM}} &= E_{gb} \cdot f_s + (E_c + E_t + E_s) \cdot f_{c} + P_{idle}\,.
\end{align}
In Equations \ref{eq:power1}, \ref{eq:power2} and \ref{eq:power3}, the following operators are used:
    \begin{itemize}
    \item $f_s$ is the sampling frequency,
    \item $f_{c}$ frequency of convolutional neural network activation,
    \item $f_{b}$ is the basic balance data sanding frequency,
    \item $\alpha^{-1}$ is the number of samples inserted in a BLE package,
    \item $P_{idle}$ power consumption of the platform in idle state, depends on the system frequency.
\end{itemize}
Figure \ref{fig:pie} shows graphically the contribution of each task to the power consumption of each OM.
\section{Conclusion}
\label{sec:conclusion}
We defined a hardware/software template for the development of a dynamically manageable sensory node, which was addressed to perform in-place analysis of sensed data. Its implementation has been tested on a low-power platform capable of recognizing simple movements on a wobble board using CNN-based data analysis. The device can reconfigure itself based on the operating modes and workload that are required. The ADAM component, which can manage device reconfiguration, contributes significantly to energy savings. On a custom dataset, a quantized neural network achieves an accuracy value greater than 97\%. By activating in-place analysis and managing the device's hardware and software components, we were able to save up to 60\% on energy. This work demonstrates the feasibility of increasing battery lifetime with near-sensor processing while also emphasizing the significance of data-dependent runtime architecture management.


\bibliographystyle{unsrt}  
\bibliography{references}

\end{document}